\pdfoutput=1
\documentclass[runningheads]{llncs}
\usepackage{amsmath}
\DeclareMathOperator*{\argmax}{argmax}
\usepackage{amssymb}
\usepackage{graphicx}
\usepackage{float}
\usepackage{array}
\usepackage{bmpsize}
\usepackage{multirow}
\usepackage{wrapfig}
\usepackage{lipsum}
\usepackage{makecell}
\usepackage{hyperref}
\hypersetup{
    colorlinks=true,
    linkcolor=blue,
    filecolor=blue,      
    urlcolor=blue,
}
\usepackage[misc]{ifsym}
\usepackage{subfigure}

\usepackage{color}

\let\oldbibliography\thebibliography
\renewcommand{\thebibliography}[1]{%
  \oldbibliography{#1}%
  \setlength{\itemsep}{0pt}%
}
\begin{document}
\title{CIA-Net: Robust Nuclei Instance Segmentation with Contour-aware Information Aggregation}

\author{Yanning Zhou$^{1}$\textsuperscript{(\Letter)}, Omer Fahri Onder$^{2}$, Qi Dou$^{3}$, Efstratios Tsougenis$^{2}$, \\Hao Chen$^{1,2}$\textsuperscript{(\Letter)}  and Pheng-Ann Heng$^{1,4}$}

\institute{$^{1}$Department of Computer Science and Engineering,
\\ The Chinese University of Hong Kong, Hong Kong SAR, China
\\\email{\{ynzhou,hchen\}@cse.cuhk.edu.hk}
\\$^{2}$Imsight Medical Technology, Co., Ltd. Hong Kong SAR, China
\\$^{3}$ Department of Computing, Imperial College London, London, UK
\\$^{4}$ Department of Computer Science and Engineering, The Chinese University of Hong Kong and Guangdong Provincial Key Laboratory of Computer Vision and Virtual Reality Technology, Shenzhen Institutes of Advanced Technology, 
\\Chinese Academy of Sciences, Shenzhen, China}

\authorrunning{Y. Zhou et al.}
\titlerunning{CIA-Net}
\makeatletter
  \newcommand\figcaption{\def\@captype{figure}\caption}
  \newcommand\tabcaption{\def\@captype{table}\caption}
\makeatother
\maketitle              
\begin{abstract}

Accurate segmenting nuclei instances is a crucial step in computer-aided image analysis to extract rich features for cellular estimation and following diagnosis as well as treatment.
While it still remains challenging because the wide existence of nuclei clusters, along with the large morphological variances among different organs make nuclei instance segmentation susceptible to over-/under-segmentation.
Additionally, the inevitably subjective annotating and mislabeling prevent the network learning from reliable samples and eventually reduce the generalization capability for robustly segmenting unseen organ nuclei.
To address these issues, we propose a novel deep neural network, namely Contour-aware Informative Aggregation Network (CIA-Net) with multi-level information aggregation module between two task-specific decoders.
Rather than independent decoders, it leverages the merit of spatial and texture dependencies between nuclei and contour by bi-directionally aggregating task-specific features.
Furthermore, we proposed a novel smooth truncated loss that modulates losses to reduce the perturbation from outliers.
Consequently, the network can focus on learning from reliable and informative samples, which inherently improves the generalization capability.
Experiments on the 2018 MICCAI challenge of Multi-Organ-Nuclei-Segmentation validated the effectiveness of our proposed method, surpassing all the other 35 competitive teams by a significant margin. 

\end{abstract}
\section{Introduction}

Digital pathology is nowadays playing a crucial role for accurate cellular estimation and prognosis of cancer~\cite{pantanowitz2010digital}.
Specifically, nuclei instance segmentation which not only captures location and density information but also rich morphology features, such as magnitude and the cytoplasmic ratio, is critical in tumor diagnosis and following treatment procedures~\cite{veta2012prognostic}.
However, automatically segmenting the nuclei at instance-level still remains challenging due to several reasons.
First, the vast existence of nuclei occlusions and clusters can easily cause over or under-segmentation, which impedes accurate morphological measurements of nuclei instances.
Second, the blurred border and inconsistent staining makes images inevitable to contain indistinguishable instances, and hence introduces subjective annotations and mislabeling, which is challenging to get robust and objective results~\cite{irshad2014crowdsourcing}.
Third, the variability in cell appearance, magnitude, and density among diverse cell types and organs requires the method to possess good generalization ability for robust analysis.

Most of the earlier methods are based on thresholding and morphological operations~\cite{cheng2009segmentation,jung2010segmenting}, which fail to find reliable threshold in the complex background.
While deep learning-based methods are generally more robust and have become the benchmark for medical image segmentation~\cite{ronneberger2015u,yi2018generative,kazeminia2018gans}.
For example, Chen et al.~\cite{chen2017dcan} proposed a deep contour-aware network (DCAN) for the task of instance segmentation that firstly harnesses the complementary information of contour and instances to separate the attached objects.
In order to utilize contour-specific features to assist nuclei prediction, BES-Net~\cite{oda2018besnet} directly concatenates the output contour features with nuclei features in decoders.
However, it only learns complementary information in nuclei branch but ignores the potentially reversed benefits from nuclei to contour, which is more essential since contour appearance is more complicated and has larger intra-variance than that of nuclei.

Another challenge is to eliminate the effect from inevitably noisy and subjective annotations. Different training strategies and loss functions have been proposed~\cite{jiang2018mentornet,xue2019robust,Goldberger2017TrainingDN,patrini2017making}.
A bootstrapped loss \cite{reed2014training} was proposed to rebalance the loss weight by taking the consistency between the label and reliable output into account.
However, when dealing with noise labeling especially the mislabeling nuclei, the network tends to predict probability with a high confidence score, where the negative log-likelihood magnitude is non-trivial and cannot be appropriately adjusted by the consistent term.
As we will show later (Sec.~\ref{subsec:loss}), these outliers overwhelm others in loss calculation and dominate the gradient.

To address the issues mentioned above, we have following contributions in this paper.
\textbf{1).} We propose an Information Aggregation Module (IAM) which enables the decoders to collaboratively refine details of nuclei and contour by leveraging the spatial and texture dependencies in bi-directionally feature aggregation.
\textbf{2).} A novel smooth truncated Loss is proposed to modulate the outliers' perturbation in loss calculation, which endows the network with the ability to robustly segment nuclei instances by focusing on learning informative samples. Moreover, eliminating outliers alleviates the network from overfitting on these noisy samples, eventually enabling the network with better generalization capability.
\textbf{3).} We validate the effectiveness of our proposed Contour-aware Information Aggregation Network (CIA-Net) with the advantages of pyramidal information aggregation and robustness on Multi-Organ Nuclei Segmentation (MoNuSeg) dataset with seven different organs, and achieved the 1st place on 2018 MICCAI Challenge, demonstrating the superior performance of the proposed approach.

\begin{figure}[H]
\center
	\includegraphics[width=0.75\textwidth]{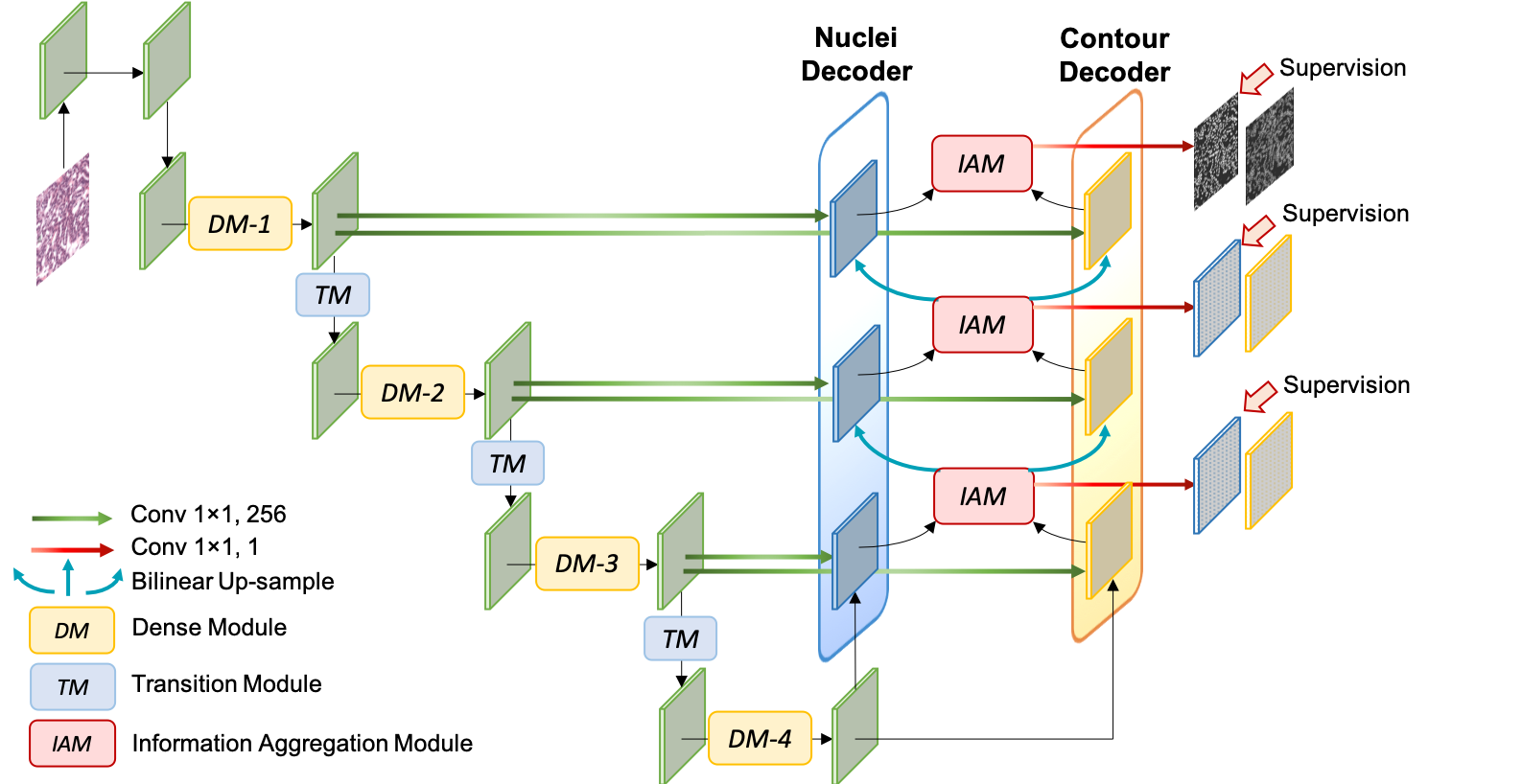}
	\caption{An overview of our proposed CIA-Net for nuclei instance segmentation.} \label{fig1}

\end{figure}

\section{Method}
Fig.~\ref{fig1} presents overview of the CIA-Net, which is a fully convolutional network (FCN) consisting of one densely connective encoder and two task-specific information aggregated decoders for refinement.
To fully leverage the benefit of complementary information from highly correlated tasks, instead of directly concatenating task-specific features, our method conducts a hierarchical refinement procedure by aggregating multi-level task-specific features between decoders.

\subsection{Densely Connected Encoder with Pyramidal Feature Extraction}
\label{subsec:encoder}
To effectively train the deep FCN, dense connectivity is introduced in encoder~\cite{huang2017densely}.
In each Dense Module (DM), let $x_{i}$ denotes the output of the $i$-th layer, dense connectivity can be described as $x_{i} = \mathcal{F}_{i}([x_{1}, x_{2}, \dots ,x_{i-1}], \mathcal{W}_{i})$.
It sets up direct connections from any bottleneck layer to all subsequent layers by concatenation, which not only effectively and efficiently reuses features but also benefits gradient back-propagation in the deep network.
Transition Module (TM) is added after DM to reduce the spatial resolution and make the features more compact, which contains a $1 \times 1$ convolution layer and an average pooling layer with a stride of 2.
Next, we hierarchically stack four DMs where each followed by a TM except the last one. For each DM, it consists of $\{6,12,24,16\}$ bottleneck layers, respectively. 

\begin{figure}[t]
\center
\includegraphics[width=0.8\textwidth]{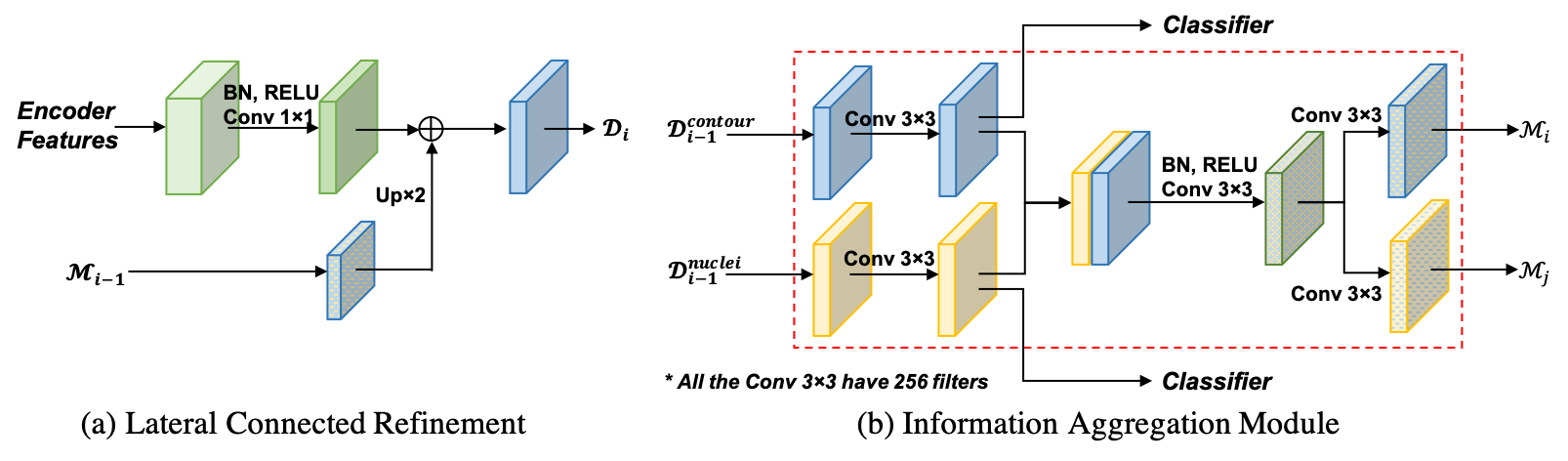}

\caption{Detail structure of (a) Lateral Connected Refinement and (b) Information Aggregation Module in proposed CIA-Net.} \label{fig2}
\end{figure}

Inspired by feature pyramid network~\cite{lin2017feature} which takes advantage of multi-scale features for accurate object detection, we propose to make full use of pyramidal features hierarchically by building multi-level lateral connections between encoder and decoders. In this way, localization and texture information from earlier layers can help the low-resolution while strong-semantic features refine the details.
The encoder features with $\{ 1/2, 1/4, 1/8\}$ of original size are passed through the lateral connections by a $1 \times 1$ convolution to reduce feature map number and merged with the upsampled deeper features in decoders by summation operation, as shown in Fig.~\ref{fig2}(a).

\subsection{Bi-directional Feature Aggregation for Accurate Segmentation}
\label{subsec:decoder}

Given that contour region encases the corresponding nuclei, it is intuitive that nuclei and contour have high spatial and contextual relevance, which is helpful for decoders to localize and focus on learning informative patterns.
In other words, the neural response from the specific kernel in nuclei branch can be considered as an extra spatial or contexture cue for localizing contour to refine details and vice versa.
In this regard, we proposed Information Aggregation Module (IAM) which aims at utilizing information from highly-correlated sub-tasks to bidirectionally aggregate the task-specific features between two decoders. Fig.~\ref{fig2}(b) shows the details of IAM structure, it takes features after lateral connection as inputs, and then selects and aggregates informative features for each sub-task.

To start the iteration, we attach a $3 \times 3$ convolution on the top of the encoder to generate the coarsest feature maps. For each decoder, the feature maps $\mathcal{M}_{i-1}$ from a higher level are upsampled by bilinear interpolation to double the resolution and added with high-resolution feature maps from encoder through lateral connections (see Fig.~\ref{fig2}(a)).
After that, the IAM takes the merged maps $\mathcal{D}_{i-1}$ as inputs and adds a $3 \times 3$ convolution without nonlinear activation to smoothen and eliminate the grid effects.
Then the smooth features are fed into the classifier to predict multi-resolution score maps.
Meanwhile, these task-specific features are concatenated along the channel dimension and then passed through two parallel convolution layers to select and integrate the complementary informative features $\mathcal{M}_{i}$ for further details refinement in the next iteration.

Besides, to prevent the network from relying on single level discriminative features, deep supervision mechanism~\cite{dou20173d} is introduced at each stage to strengthen learning of multi-level contextual information. This also benefits training of deeper network architectures by shortening the back-propagation path.

\subsection{Smooth Truncated Loss for Robust Nuclei Segmentation}
\label{subsec:loss}
\begin{figure}[t]

\centering
\includegraphics[width=0.7\textwidth]{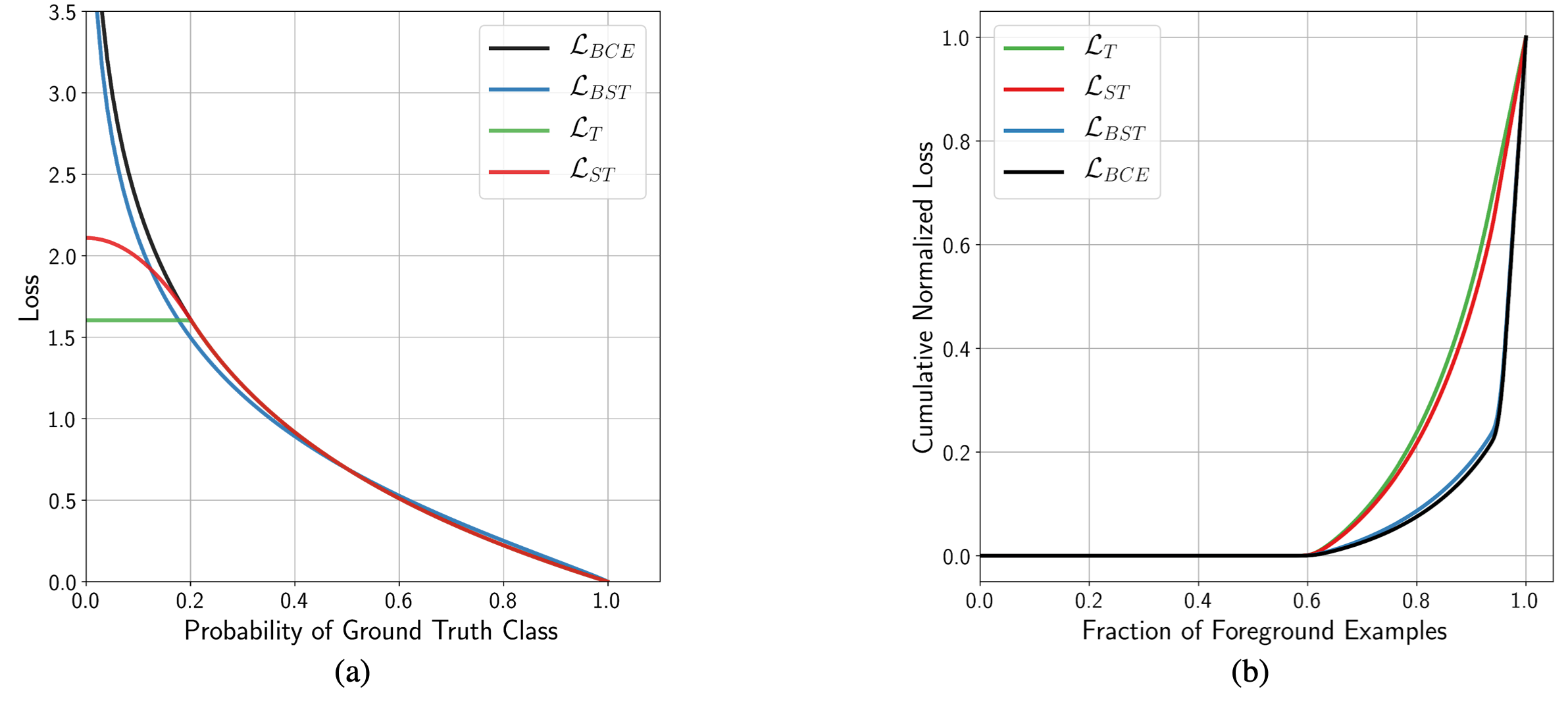}

\caption{Visualization of different loss functions (a) with $\gamma=0.2$ and the cumulative loss functions of normalized loss from foreground regions (b).} \label{cdf}

\end{figure}

The existence of blurred edge and inconsistent staining makes images inevitably contain indistinguishable instances, which leads to subjective annotations such as mislabelled objects and inaccurate boundary.
Additionally, to enhance the ability to split attached nuclei, conventional practice is to preprocess the training ground truth by subtracting the dilated contour mask, which is also suboptimal and has the risk of introducing noises.
Both factors show that it is unavoidable for pixel-wise nuclei annotations to contain imperfect labels, which is harmful to network training from at least two aspects.
Firstly, the inaccurate labeling encountered during training has the tendency to overwhelm other regions in loss calculation and dominate the gradients.
This phenomenon is observed from the sorted cumulative distribution function of normalized loss in Fig.~\ref{cdf}(b) using a converged model.
Notice that top $10\%$ samples account for more than $80\%$ value of cross-entropy loss, which prevents network learning from informative samples during gradient back-propagation.
Secondly, forcibly learning the subjective labeling would eventually push the network to particularly fit them and tend to overfitting, which is even more pernicious when predicting unseen organ nuclei.
To handle the noisy and incomplete labeling,~\cite{reed2014training} proposed bootstrapped loss ($\mathcal{L}_{BST}$) to rebalance the loss weight by considering the consistency between the label and reliable output.
However, as can be seen in Fig.~\ref{cdf}(b), when faced with errors with low predicted probability, it cannot easily compensate for the loss with non-trivial magnitude.

To solve this problem, our insight is to reduce outliers' interference in training by modulating contribution in loss calculation.
Under the premise of high credibility of network prediction, the majority of outliers will lie in low predicted probability regions and get large values of error.
Inspired by Huber loss~\cite{friedman2001elements} for robust regression, which is quadratic for small values of error and linear for large values to decline the influence of outliers, we propose the prototype of loss function, namely Truncated Loss ($\mathcal{L}_{T}$), which reduces the contribution of outliers with high confidence prediction.
Let $p_{t}$ denotes the predicted probability of the ground truth, $p_{t} = p$ if $t=1$ and $p_{t} =1 - p$ otherwise, in which $t\in\{0, 1\}$ specifics the ground truth label.
Formally, the loss is truncated when the corresponding $p_{t}$ is smaller than a threshold $\gamma \in[0, 0.5]$:
\begin{equation}
 \mathcal{L}_{T} = -\max (\log(p_{t}), \log(\gamma)).
 \label{eq:l_t}
\end{equation}
The truncated loss only clips outliers with $p_{t}<\gamma$, while preserves loss value for the other.
Intuitively, this operation adds a constraint of maximum contribution in loss calculation from each pixel and hence can ease the gradient domination from outliers and benefit of learning the informative samples.
However, in Eq.~(\ref{eq:l_t}) the derivative of $\mathcal{L}_{T}$ at clipping point $p_{t}=\gamma$ is undefined.
Meanwhile, the perturbation of low $p_{t}$ prediction will not be reflected in loss calculation if we force the loss value larger than the threshold to a constant, therefore the smoothed version is preferred for optimization.
In this regard, we propose Smooth Truncated Loss $\mathcal{L}_{ST}$:

\begin{equation}
\mathcal{L}_{ST} = \left\{
\begin{array}{lr}
- \log(\gamma) + \frac{1}{2} (1 -\frac{p_{t}^{2}}{\gamma^{2}})  , & p_{t} <\gamma \\
- \log(p_{t}), &  p_{t}\geqslant\gamma
\end{array}
\right.
\label{eq:l_st}
\end{equation}

A quadratic function with the same value and derivative as negative log-likelihood at the truncated point $\gamma$ is used to modulate the loss weight for outliers.
By incorporating constraint for the loss magnitude, it reduces the contribution of outliers, where the smaller $p_{t}$, the more considerable modulation. This, in turn, let the network discard the indistinguishable parts and focus on informative and learnable regions.
Furthermore, by reducing the influence of the outlier samples that interferences the network training, it encourages the network to predict with higher confidence scores and narrow the uncertain regions, which is helpful for alleviating over-segmentation.

\subsection{Overall Loss Function}
\label{subsec:details}
Based on the proposed Smooth Truncated Loss, we can derive the overall loss function.
Note that the contour prediction is much more difficult than that of nuclei due to irregularly curved form.
In this case, the primary component of regions with high loss is not by the outliers, but the inlier samples, and hence utilizing truncated loss may confuse the network. Instead, we use Soft Dice Loss to learn the shape similarity:
\begin{equation}
\mathcal{L}_{Dice} = 1 - \frac{2\sum_{i=1}^{n}p_{i}q_{i}}{\sum_{i=1}^{n}p_{i}^{2} + \sum_{i=1}^{n}q_{i}^{2}},
\end{equation}
where $p_{i}$ denotes the predicted probability of i-th pixel and $q_{i}$ denotes the corresponding ground truth.
In sum, the total loss function for proposed CIA-Net training is:
\begin{equation}
\mathcal{L}_{total} =  \mathcal{L}_{ST} + \lambda \mathcal{L}_{Dice} +\beta\|\mathcal{W}\|^{2}_{2},
\end{equation}
where the first and second terms calculate error from contour and nuclei prediction respectively, and the third term is the weight decay.
$\lambda$ and $\beta$ are hyper-parameters to balance three components.

\section{Experimental Results}
\subsection{Dataset and Evaluation Metrics}
We validated our proposed method on MoNuSeg dataset of 2018 MICCAI challenge, which contains 30 images (size: $1000 \times 1000$ ) captured by The Cancer Genomic Atlas (TCGA) from whole slide images (WSIs)~\cite{kumar2017dataset}.
The dataset consists of breast, liver, kidney, prostate, bladder, colon and stomach containing both benign and malignant cases, which is then divided into training set (\textit{Train}), test set1 from the same organs of training data (\textit{Test1}) and test set2 from unseen organs (\textit{Test2}) with 14, 8 and 6 images, respectively.
The \textit{Train} contains 4 organs - breast, kidney, liver and prostate with 4 images from each organ, the \textit{Test1} includes 2 images from per organ mentioned in \textit{Train}, and \textit{Test2} contains 2 images from each unseen organ, i.e., bladder, colon and stomach.

We employed Average Jaccard Index (AJI)~\cite{kumar2017dataset} for comparison, which considers an aggregated intersection cardinality numerator and an aggregated union cardinality denominator for all ground truth and segmented nuclei.
Let $G = \{G_{1}, G_{2},\dots G_{n}\}$ denotes the set of instance ground truths, $S= \{S_{1}, S_{2},\dots S_{m}\}$ denotes the set of segmented objects and $N$ denotes the set of segmented objects with none intersection to ground truth. AJI = $\displaystyle\frac{\sum_{i=1}^{n}G_{i}\bigcap S_{j}}{\sum_{i=1}^{n}G_{i}\bigcup S_{j}+\sum_{k\in N}S_{k}}$, where $j = \displaystyle\argmax_{k} \frac{G_{i}\bigcap S_{k}}{G_{i}\bigcup  S_{k}}$.
F1-score ($\displaystyle F1 =\frac{2\cdot Precision\cdot Recall}{Precision+Recall}$)~\cite{chen2017dcan} is used for nuclei instance detection performance evaluation and we also report it for reference.

\subsection{Implementation Details}
We implemented our network using Tensorflow (version 1.7.0).
The default parameters provided at \href{https://github.com/pudae/tensorflow-densenet}{https://github.com/pudae/tensorflow-densenet} is used in the Densenet backbone.
Stain normalization method~\cite{macenko2009method} was performed before training.
Data augmentations including crop, flip, elastic transformation and color jitter were utilized.
The outputs of nuclei and contour maps were first subtracted and then the connected components were detected get the final results.
The network was trained on one NVIDIA TITAN Xp GPU with a mini-batch size of three.  
We utilized the pre-trained DenseNet model~\cite{huang2017densely} from ImageNet to initialize the encoder.
The hyper-parameters $\lambda$ and $\beta$ were set as 0.42 and 0.0001 to balance the loss and regularization.
AdamW optimizer was used to optimize the whole network and learning rate was initialized as 0.001 and decayed according to cosine annealing and warm restarts strategy~\cite{loshchilov2017fixing}.

\subsection{Evaluation and Comparison}
\textbf{Effectiveness of contour-aware information aggregation architecture.}
Firstly, we conduct a series of experiments to compare different informative feature aggregation strategies in decoders:
(1) \textit{Cell Profiler}~\cite{carpenter2006cellprofiler}: a python-based software for computational pathology employing intensity thresholoding method.
(2) \textit{Fiji}~\cite{schindelin2012fiji}: a Java-based software utilizing watershed transform nuclear segmentation method.
(3) \textit{CNN3}~\cite{kumar2017dataset}: a 3-class FCN without deep dense connectivity.
(4) \textit{DCAN}~\cite{chen2017dcan}: a deep FCN with multi-task learning strategy for objects and contours.
(5) \textit{PA-Net}~\cite{liu2018path}: a modified path aggregation network by adding path augmentation in two independent decoders to enhance the instance segmentation performance.
(6) \textit{BES-Net}~\cite{oda2018besnet}: the original boundary-enhanced segmentation network which concatenated contour features with nuclei features to enhance learning in boundary region.
(7) \textit{CIA-Net w/o IAM}: the proposed network architecture with two independent decoders for nuclei and contour prediction respectively, but without Information Aggregation Module in decoders.
(8) \textit{Proposed CIA-Net}: Our Contour-aware Information Aggregation Network with Information Aggregation Module between nuclei and contour decoders.
Notice that unless specified otherwise, we utilized the same encoder structure with pyramidal feature extraction strategy and loss functions to establish fair comparison.
\begin{table}
\centering
\caption{Performance comparison of different methods on \textit{Test1} (seen organ) and \textit{Test2} (unseen organ). }
\begin{tabular}{p{0.55in}<{\centering} p{1.5in}<{\centering}|p{0.55in}<{\centering}|p{0.55in}<{\centering}|p{0.55in}<{\centering}|p{0.55in}<{\centering}}
 \hline
  &\multirow{3}{*}{Method} & \multicolumn{2}{c|}{AJI} &  \multicolumn{2}{c}{F1-score} \\
  \cline{3-6}
  & & Test1	& Test2& Test1 & Test2	\\
   \hline
  (1)& Cell Profiler~\cite{carpenter2006cellprofiler}& 0.1549	&0.0809 	 & 0.4143&0.3917  \\
  \hline
  (2) &Fiji~\cite{schindelin2012fiji}& 0.2508	&0.3030 	  & 0.6402 &0.6978  \\
  \hline
  (3)&CNN3~\cite{kumar2017dataset}& 0.5154	&0.4989 	 & 0.8226 &0.8322 \\
  \hline
  (4)&DCAN~\cite{chen2017dcan}&0.6082 	&	0.5449	 		& 0.8265&0.8214\\
  \hline
  (5)&PA-Net~\cite{liu2018path}& 0.6011 	&0.5608&0.8156&0.8336 \\
  \hline
  (6)&BES-Net~\cite{oda2018besnet}&	0.5906	 &0.5823		 &0.8118&0.7952\\
   \hline
   \hline
   (7)  &CIA-Net w/o IAM& 0.6106	&0.5817		 &\textbf{0.8279} &0.8356\\
  \hline

  (8)  & \textbf{Proposed CIA-Net}& \textbf{0.6129}& \textbf{0.6306}	 & 0.8244	& \textbf{0.8458}\\

   \hline
\end{tabular}

\end{table}

It is observed that all CNN-based approaches achieved much higher results on all evaluation criterions than conventional approaches, highlighting the superiority of deep learning based methods for segmentation related tasks.
Moreover, results from (4) to (8) have a striking improvement regarding the evaluation metric of AJI on both \textit{Test1} and \textit{Test2} compared with (3), validating the efficacy of dense connectivity structure, which is more powerful to leverage multilevel features and mitigate gradient vanishing in training deep neural network.
While methods (4) to (7) achieved comparable performance on the evaluation performance of \textit{Test1}, the
results from \textit{BES-Net} and proposed \textit{CIA-Net w/o IAM} outperform others significantly on AJI of \textit{Test2}, demonstrating the exploitation of high spatial and context relevance between nuclei and contour can generate task-specific features for assisting feature refinement between both tasks. This can help enhance the generalization capability to unseen data. 
Meanwhile, in comparison with \textit{BES-Net} and proposed \textit{CIA-Net w/o IAM}, our proposed network  \textit{CIA-Net} further outperforms these two methods consistently regarding the metric of AJI, achieving overall best performance and boosting results to 0.6306 on \textit{Test2} and 0.6129 on \textit{Test1}.
Different from \textit{BES-Net} which directly concatenates features in contour decoder to nuclei branch, the proposed \textit{CIA-Net} with IAM bi-directionally aggregating the task-specific features and passing them through parallel convolutions to iteratively aggregate informative features in decoders.
Therefore, it is a learnable procedure for network to find favorable features, which mutually benefits two sub-tasks. 
Compared with the improvement on AJI, the improvement on F1-score is less significant, this is because AJI is a segment-based metric while F1-score is the detection-based metric. 
\\
\textbf{Effectiveness of proposed Smooth Truncated loss.}
Toward the potential of clinical application, the proposed method should be robust under the numerous circumstances, especially for the diffused-chromatin and attached nuclei in unseen organs, which is evaluated in \textit{Test2} set.
We compare the results of our proposed CIA-Net with four different functions:
(1) $\mathcal{L}_{BCE}$: Binary Cross-Entropy loss.
(2) $\mathcal{L}_{BST}$: Soft Bootstrapped loss by rebalancing the loss weight.
(3) $\mathcal{L}_{T}$: Proposed Truncated loss without smoothing around truncated point, i.e., Eq.~(\ref{eq:l_t}).
(4) $\mathcal{L}_{ST}$: Proposed Smooth Truncated loss which utilizes quadratic function as soft modulation, i.e., Eq.~(\ref{eq:l_st}).
\begin{figure}[htb]

  \begin{minipage}[b]{0.55\textwidth}
    \centering
    \renewcommand\arraystretch{1.1}
\begin{tabular}{p{0.48in}<{\centering}|p{0.48in}<{\centering}|p{0.48in}<{\centering}|p{0.48in}<{\centering}|p{0.48in}<{\centering}}
  \hline
 \multirow{2}{*}{Loss} & \multicolumn{2}{c|}{AJI} & \multicolumn{2}{c}{F1-score} \\
   \cline{2-5}
  & Test1	& Test2	& Test1& Test2\\
 \hline
  $\mathcal{L}_{BCE}$	 &0.6104	&0.5934	 &0.8303	&0.8433	\\

 \hline
  $\mathcal{L}_{BST}$&0.6123&0.6058&\textbf{0.8415}&0.8260\\
 \hline
  $\mathcal{L}_{T} $&  0.6133	& 0.6153 	 &0.8377 & 0.8307	\\
 \hline
  $\mathcal{L}_{ST} $	& 0.6129& \textbf{0.6306} & 0.8244	& \textbf{0.8458}	\\
 \hline
\end{tabular}
    \tabcaption{Comparison of proposed CIA-Net \protect\\with different loss functions.}
    \label{loss}
  \end{minipage}
  \begin{minipage}[b]{0.45\textwidth}
    \centering
    \includegraphics[width=0.8\textwidth]{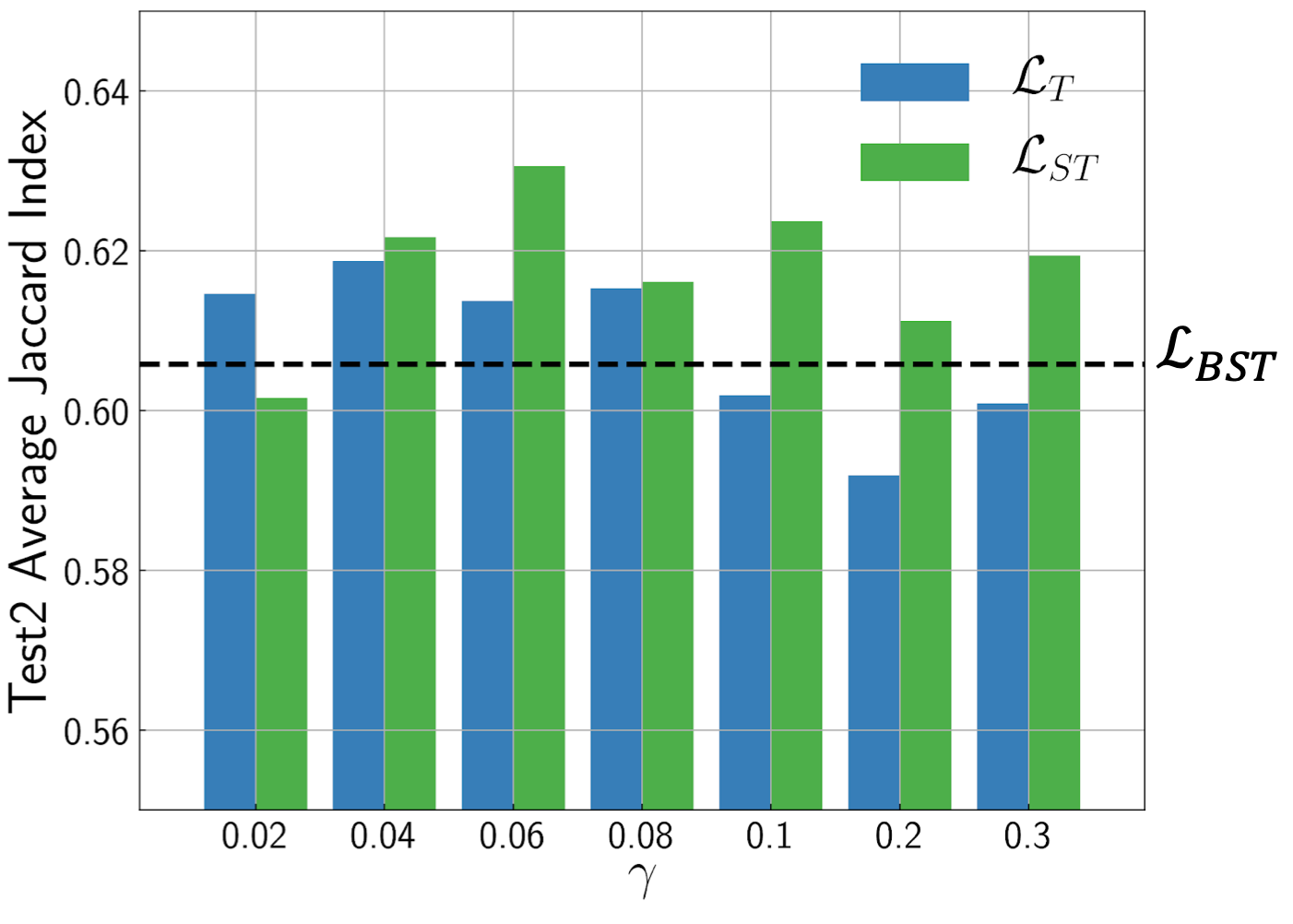}
    \vspace{-14pt}
    \caption{Results of varying $\gamma$ for $\mathcal{L}_{T}$ and $\mathcal{L}_{ST}$ on \textit{Test2}.}
    \label{gamma}
  \end{minipage}%
\end{figure}

As can be seen in Table~\ref{loss}, the improvement of $\mathcal{L}_{BST}$ compared to $\mathcal{L}_{BCE}$ is limited.
Compared with first two rows, results from $\mathcal{L}_{T}$ and  $\mathcal{L}_{ST}$ outperform others on \textit{Test2} consisting of unseen organs by a large margin (nearly $2.5\%$ for $\mathcal{L}_{ST}$ and $1\%$ for $\mathcal{L}_{T}$) regarding the metric of AJI, and are analogous on \textit{Test1}. 
The proposed $\mathcal{L}_{ST}$ achieved significant improvements in comparison with $\mathcal{L}_{T}$ on \textit{Test2}, shows it is less sensitive on $\gamma$ and has better generalization capability on different organ images. 
The proposed Smooth Truncated loss introduces one new hyper-parameter, the truncating parameter $\gamma$, which controls the starting point of down-weighting outliers.
When $\gamma=0$, the loss function degenerates into Binary Cross-entropy $\mathcal{L}_{BCE}$ .
As $\gamma$ increases, more examples with $p_{t}$ lower than $\gamma$ are considered as outliers or less informative samples to down-weight in loss calculation.
Fig.~\ref{gamma} illustrates the influence of varying $\gamma$. 
We can see $\mathcal{L}_{ST}$ have a striking overall improvement compared with $\mathcal{L}_{BST}$ and $\mathcal{L}_{T}$. More importantly, results from $\mathcal{L}_{ST}$ demonstrate less sensitivity for choosing different $\gamma$.
\begin{figure}[h]
\center
	\includegraphics[width=0.75\textwidth]{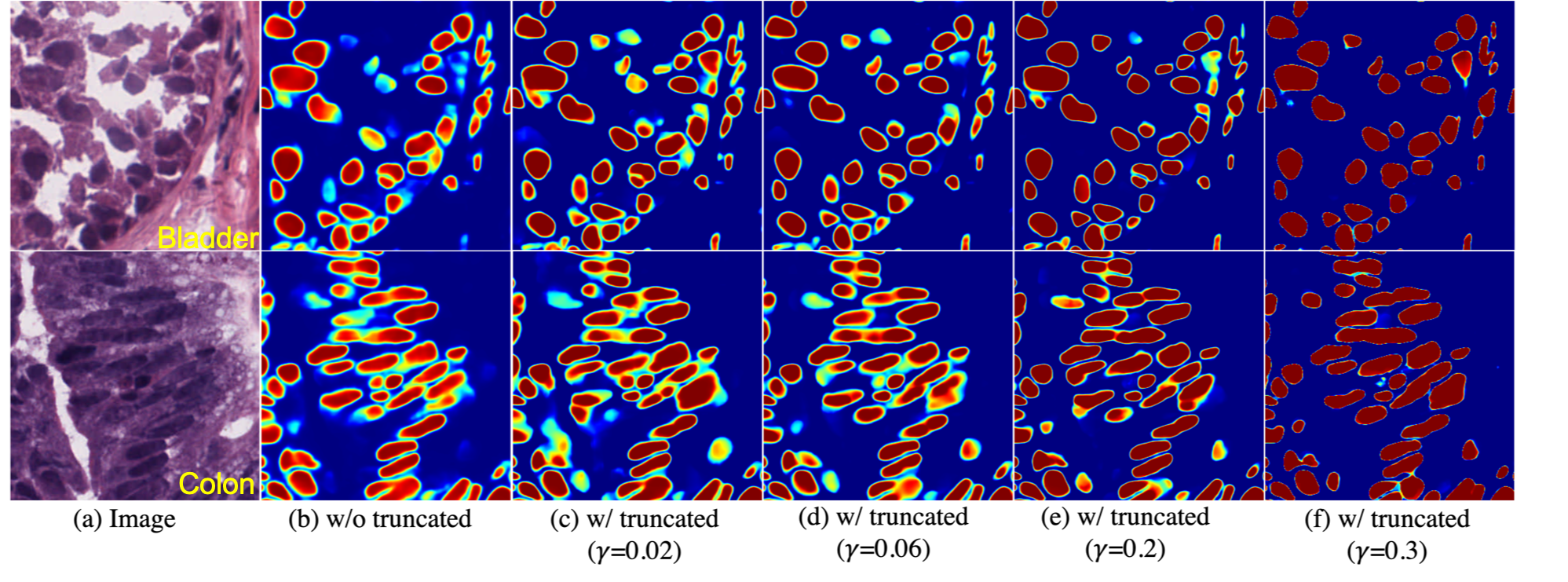}

	\caption{Visualization of heatmaps in different $\gamma$ values from $Test2$.}
	\label{heatmap}
\end{figure}

\begin{figure}[!h]
\centering
	\includegraphics[width=0.8\textwidth]{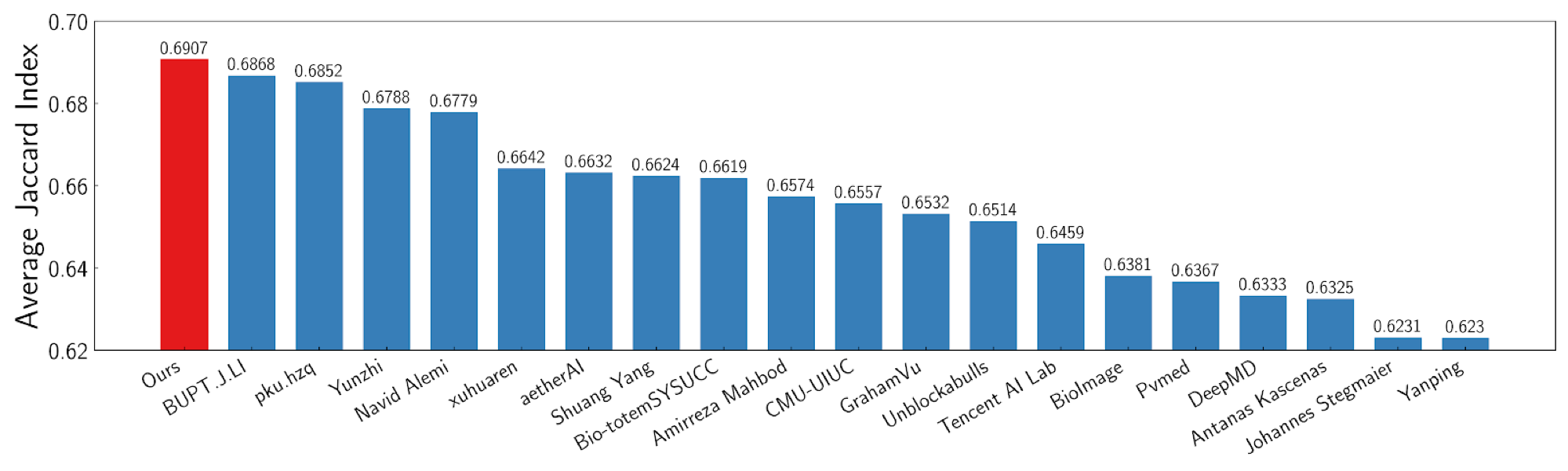}

	\caption{ The instance segmentation results of different methods in 2018 MICCAI Multi-Organ Nuclei Segmentation Challenge (top 20 of 36 methods are shown in figure). }
		\label{test}

\end{figure}

We visualize the nuclei heatmaps from setting different $\gamma$ in $\mathcal{L}_{ST}$ (see Fig.~\ref{heatmap}) to give an intuitive understanding of our proposed method.
It is observed that heatmaps trained by $\mathcal{L}_{BCE}$ (Fig.~\ref{heatmap}(b)) contain massive blur and noise, which is unfavorable for binarizing instances.
As $\gamma$ increases, the heatmaps turn to be more concrete with less uncertain areas, which is of great significance for instance segmentation to prevent over-segmentation.
While setting too large $\gamma$ increases the risk of under-segmentation, as can be seen in Fig.~\ref{heatmap}(f).
This is because over suppressing low $p_{t}$ region also penalties learning from informative inlier samples, especially boundary regions where the $p_{t}$ is relatively small.
\\
\textbf{2018 MICCAI MoNuSeg Challenge results.}
We employed above entire dataset for training and 14 additional images provided by organizer for independent evaluation with ground truth held out\footnote{\href{https://monuseg.grand-challenge.org}{https://monuseg.grand-challenge.org}}.
Top 20 results of 36 teams are shown in Fig.~\ref{test}.
Our submitted entry surpassed all the other methods, highlighting the strength of the proposed CIA-Net and Smooth Truncated loss.

\begin{figure}[!h]
\centering
	\includegraphics[width=0.75\textwidth]{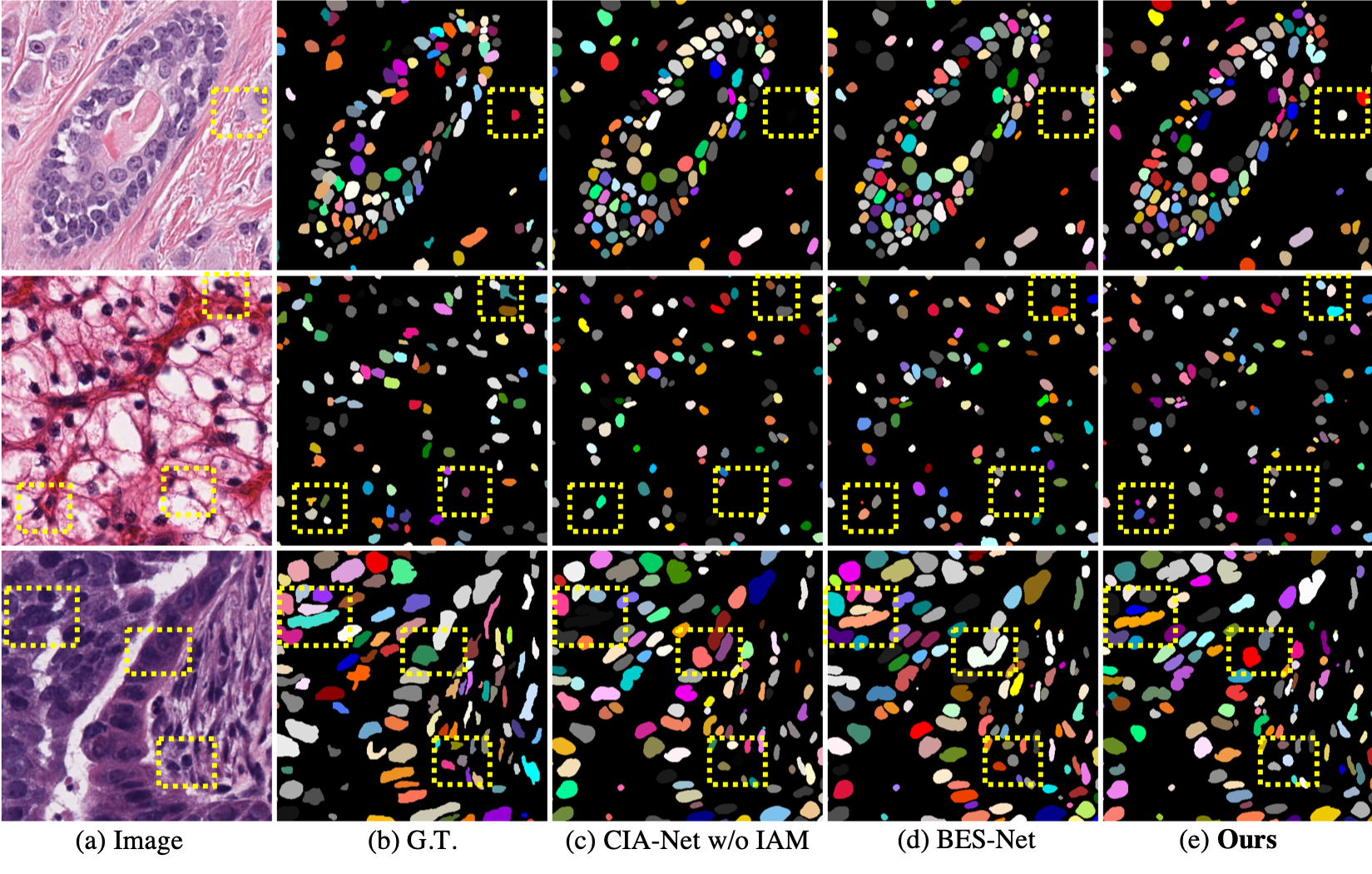}
	\caption{Qualitative results of multi-organ nuclei (from top to bottom: breast, kidney, colon) on \textit{Test1} and \textit{Test2}. Yellow rectangles highlight the difference among predictions. }
		\label{instance_results}
\end{figure}

\textbf{Qualitative analysis.}
Fig.~\ref{instance_results} shows representative samples from \textit{Test1} and \textit{Test2} with challenging cases such as diffuse-chromatin nuclei and irregular shape.
Notice that our proposed \textit{CIA-Net} (Fig.~\ref{instance_results}(e)) can generate the segmentation results similar to the annotations of human experts, outperforming others with less over or under-segmentation on the prolific nuclei clusters and attached cases.

\section{Conclusion}
Instance-level nuclei segmentation is the pivotal step for cell estimation and further pathological analysis.
In this paper, we propose CIA-Net with the smooth truncated loss to tackle the challenges of prolific nuclei clusters and inevitable labeling noise in pathological images.
Our method inherently can be adapted to a wide range of medical image segmentation tasks to boost the performance such as histology gland segmentation.
\\
\\
\textbf{Acknowledgments.} This work was supported by Hong Kong Innovation and Technology Fund (Project No. ITS/041/16), Guangdong province science and technology plan project (No.2016A020220013).
 \bibliographystyle{splncs04}

\begin{thebibliography}{10}
\providecommand{\url}[1]{\texttt{#1}}
\providecommand{\urlprefix}{URL }
\providecommand{\doi}[1]{https://doi.org/#1}

\bibitem{carpenter2006cellprofiler}
Carpenter, A.E., Jones, T.R., Lamprecht, M.R., Clarke, C., Kang, I.H., Friman,
  O., Guertin, D.A., Chang, J.H., Lindquist, R.A., Moffat, J., et~al.:
  Cellprofiler: image analysis software for identifying and quantifying cell
  phenotypes. Genome biology  \textbf{7}(10), ~R100 (2006)

\bibitem{chen2017dcan}
Chen, H., Qi, X., Yu, L., Dou, Q., Qin, J., Heng, P.A.: Dcan: Deep
  contour-aware networks for object instance segmentation from histology
  images. Medical image analysis  \textbf{36},  135--146 (2017)

\bibitem{cheng2009segmentation}
Cheng, J., Rajapakse, J.C., et~al.: Segmentation of clustered nuclei with shape
  markers and marking function. IEEE Trans. Biomed. Eng.  \textbf{56}(3),
  741--748 (2009)

\bibitem{dou20173d}
Dou, Q., Yu, L., Chen, H., Jin, Y., Yang, X., Qin, J., Heng, P.A.: 3d deeply
  supervised network for automated segmentation of volumetric medical images.
  Medical image analysis  \textbf{41},  40--54 (2017)

\bibitem{friedman2001elements}
Friedman, J., Hastie, T., Tibshirani, R.: The elements of statistical learning,
  vol.~1. Springer series in statistics New York (2001)

\bibitem{Goldberger2017TrainingDN}
Goldberger, J., Ben-Reuven, E.: Training deep neural-networks using a noise
  adaptation layer. In: ICLR 2017 (2017)

\bibitem{huang2017densely}
Huang, G., Liu, Z., Van Der~Maaten, L., Weinberger, K.Q.: Densely connected
  convolutional networks. In: CVPR (2017)

\bibitem{irshad2014crowdsourcing}
Irshad, H., Montaser-Kouhsari, L., Waltz, G., Bucur, O., A~Nowak, J., Dong, F.,
  Knoblauch, N., Beck, A.: Crowdsourcing image annotation for nucleus detection
  and segmentation in computational pathology: evaluating experts, automated
  methods, and the crowd. In: Pac Symp Biocomput. pp. 294--305. World
  Scientific (2014)

\bibitem{jiang2018mentornet}
Jiang, L., Zhou, Z., Leung, T., Li, L.J., Fei-Fei, L.: Mentornet: Learning
  data-driven curriculum for very deep neural networks on corrupted labels. In:
  ICML (2018)

\bibitem{jung2010segmenting}
Jung, C., Kim, C.: Segmenting clustered nuclei using h-minima transform-based
  marker extraction andcontour parameterization. IEEE Trans. Biomed. Eng.
  \textbf{57}(10),  2600--2604 (2010)

\bibitem{kazeminia2018gans}
Kazeminia, S., Baur, C., Kuijper, A., van Ginneken, B., Navab, N., Albarqouni,
  S., Mukhopadhyay, A.: Gans for medical image analysis. arXiv preprint
  arXiv:1809.06222  (2018)

\bibitem{kumar2017dataset}
Kumar, N., Verma, R., Sharma, S., Bhargava, S., Vahadane, A., Sethi, A.: A
  dataset and a technique for generalized nuclear segmentation for
  computational pathology. IEEE Trans. Med. Imaging  \textbf{36}(7),
  1550--1560 (2017)

\bibitem{lin2017feature}
Lin, T.Y., Doll{\'a}r, P., Girshick, R.B., He, K., Hariharan, B., Belongie,
  S.J.: Feature pyramid networks for object detection. In: IEEE CVPR (2017)

\bibitem{liu2018path}
Liu, S., Qi, L., Qin, H., Shi, J., Jia, J.: Path aggregation network for
  instance segmentation. In: IEEE CVPR (2018)

\bibitem{loshchilov2017fixing}
Loshchilov, I., Hutter, F.: Fixing weight decay regularization in adam. arXiv
  preprint arXiv:1711.05101  (2017)

\bibitem{macenko2009method}
Macenko, M., Niethammer, M., Marron, J.S., Borland, D., Woosley, J.T., Guan,
  X., Schmitt, C., Thomas, N.E.: A method for normalizing histology slides for
  quantitative analysis. In: IEEE ISBI (2009)

\bibitem{oda2018besnet}
Oda, H., Roth, H.R., Chiba, K., Sokoli{\'c}, J., Kitasaka, T., Oda, M., Hinoki,
  A., Uchida, H., Schnabel, J.A., Mori, K.: Besnet: Boundary-enhanced
  segmentation of cells in histopathological images. In: MICCAI 2018. LNCS,
  vol. 11071, pp. 228--236 (2018)

\bibitem{pantanowitz2010digital}
Pantanowitz, L.: Digital images and the future of digital pathology. J Pathol
  Inform  \textbf{1} (2010)

\bibitem{patrini2017making}
Patrini, G., Rozza, A., Krishna~Menon, A., Nock, R., Qu, L.: Making deep neural
  networks robust to label noise: A loss correction approach. In: IEEE CVPR
  (2017)

\bibitem{reed2014training}
Reed, S., Lee, H., Anguelov, D., Szegedy, C., Erhan, D., Rabinovich, A.:
  Training deep neural networks on noisy labels with bootstrapping. arXiv
  preprint arXiv:1412.6596  (2014)

\bibitem{ronneberger2015u}
Ronneberger, O., Fischer, P., Brox, T.: U-net: Convolutional networks for
  biomedical image segmentation. In: MICCAI 2015. LNCS, vol.~9351, pp. 234--241
  (2015)

\bibitem{schindelin2012fiji}
Schindelin, J., Arganda-Carreras, I., Frise, E., Kaynig, V., Longair, M.,
  Pietzsch, T., Preibisch, S., Rueden, C., Saalfeld, S., Schmid, B., et~al.:
  Fiji: an open-source platform for biological-image analysis. Nature methods
  \textbf{9}(7), ~676 (2012)

\bibitem{veta2012prognostic}
Veta, M., Kornegoor, R., Huisman, A., Verschuur-Maes, A.H., Viergever, M.A.,
  Pluim, J.P., Van~Diest, P.J.: Prognostic value of automatically extracted
  nuclear morphometric features in whole slide images of male breast cancer.
  Mod. Pathol.  \textbf{25}(12), ~1559 (2012)

\bibitem{xue2019robust}
Xue, C., Dou, Q., Shi, X., Chen, H., Heng, P.A.: Robust learning at noisy
  labeled medical images: Applied to skin lesion classification. In: IEEE ISBI
  (2019)

\bibitem{yi2018generative}
Yi, X., Walia, E., Babyn, P.: Generative adversarial network in medical
  imaging: A review. arXiv preprint arXiv:1809.07294  (2018)

\end{thebibliography}
 {
 
}

\end{document}